\documentclass[a4paper,11pt]{article}
\newcommand*{\VIX}{}

\usepackage[utf8]{inputenc}
\usepackage[margin=1in]{geometry}
\usepackage[colorlinks,linkcolor=blue,citecolor=blue]{hyperref}
\usepackage{float,framed}
\usepackage{algorithm,algorithmic}
\usepackage[numbers]{natbib}
\usepackage{enumitem}
\usepackage{bm}
\usepackage{xspace}
\usepackage{bbm} %for \mathbbm{1}
\usepackage{cmap} %Makes ff, fi copyable and searchable. Always use this!

\usepackage{amsmath,amsfonts,amssymb}
\usepackage{mathtools}
\usepackage{amsthm} % better to load after ams math
\usepackage{latexsym}
\usepackage{relsize}

\usepackage[capitalise]{cleveref}
\usepackage{xcolor}
\usepackage{dsfont}

\def\yhat{\hat{{y}}}

\newcommand{\ignore}[1]{}

\theoremstyle{plain}
\newtheorem{theorem}{Theorem}
\newtheorem{lemma}[theorem]{Lemma}
\newtheorem{corollary}[theorem]{Corollary}

\newtheorem*{theorem*}{Theorem}
\newtheorem*{lemma*}{Lemma}
\newtheorem*{corollary*}{Corollary}
\newtheorem*{proposition*}{Proposition}
\newtheorem*{claim*}{Claim}
\newtheorem*{fact*}{Fact}
\newtheorem*{observation*}{Observation}

\theoremstyle{definition}
\newtheorem{definition}[theorem]{Definition}

\newtheorem*{definition*}{Definition}
\newtheorem*{remark*}{Remark}
\newtheorem*{example*}{Example}

 \theoremstyle{plain}
\newtheorem*{theoremaux}{\theoremauxref}
\gdef\theoremauxref{1}

\DeclareMathAlphabet{\mathbfsf}{\encodingdefault}{\sfdefault}{bx}{n}

\DeclareMathOperator*{\argmin}{arg\,min}

\newcommand{\norm}[1]{\|#1\|}

\newcommand{\wt}[1]{\smash{\widetilde{#1}}}

\newcommand{\poly}{\mathrm{poly}}

\newcommand{\reals}{\mathbb{R}}
\newcommand{\eps}{\varepsilon}
\newcommand{\ep}{\varepsilon}

\renewcommand{\leq}{~\le~}
\renewcommand{\geq}{~\ge~}

\let\oldtfrac\tfrac
\renewcommand{\tfrac}[2]{\smash{\oldtfrac{#1}{#2}}}

\let\nablaold\nabla
\renewcommand{\nabla}{\nablaold\mkern-2.5mu}
\def\shownotes{1}  \ifnum\shownotes=1
\newcommand{\authnote}[2]{$\ll$\textsf{\footnotesize #1 notes: #2}$\gg$}
\else
\newcommand{\authnote}[2]{}
\fi

\newcommand{\R}[0]{\mathbb{R}}
\newcommand{\al}[0]{\alpha}
\newcommand{\be}[0]{\beta}

\newcommand{\de}[0]{\delta}

\newcommand{\eph}[0]{\frac{\varepsilon}{2}}

\newcommand{\La}[0]{\Lambda}

\newcommand{\rh}[0]{\rho}
\newcommand{\te}[0]{\theta}
\newcommand{\Te}[0]{\Theta}

\newcommand{\om}[0]{\omega}
\newcommand{\Om}[0]{\Omega}
\newcommand{\si}[0]{\sigma}

\newcommand{\nin}[0]{\not\in}

\newcommand{\iy}[0]{\infty}
\newcommand{\rc}[1]{\frac{1}{#1}}

\newcommand{\fc}[2]{\frac{#1}{#2}}
\newcommand{\sfc}[2]{\sqrt{\frac{#1}{#2}}}
\newcommand{\pf}[2]{\pa{\frac{#1}{#2}}}%Shortcut for fraction with parentheses

\newcommand{\ab}[1]{\left| {#1} \right|}
\newcommand{\an}[1]{\left\langle {#1}\right\rangle}

\newcommand{\bc}[1]{\left\{ {#1} \right\}}

\newcommand{\pa}[1]{\left( {#1} \right)}

\newcommand{\ve}[1]{\left\Vert {#1}\right\Vert}

\newcommand{\set}[2]{\left\{{#1}:{#2}\right\}}

\newcommand{\ol}[1]{\overline{#1}}

\newcommand{\ub}[2]{\underbrace{#1}_{#2}}

\newcommand{\diag}{\operatorname{diag}}

\newcommand{\pull}[9]{
#1\ar@/_/[ddr]_{#2} \ar@{.>}[rd]^{#3} \ar@/^/[rrd]^{#4} & &\\
& #5\ar[r]^{#6}\ar[d]^{#8} &#7\ar[d]^{#9} \\}

\newcommand{\cmp}[9]{
\xymatrix{
#1 \ar[r]^{#4}{#5} \ar@/_2pc/[rr]^{#8}_{#9} & #2 \ar[r]^{#6}_{#7} & #3
}
}

\newcommand{\ha}[1]{\ar@{^(->}[#1]}
\newcommand{\ls}[1]{\ar@{-}[#1]}
\newcommand{\sj}[1]{\ar@{->>}[#1]}
\newcommand{\aq}[1]{\ar@{=}[#1]}
\newcommand{\acir}[1]{\ar@{}[#1]|-{\textstyle{\circlearrowright}}}
\newcommand{\acil}[1]{\ar@{}[#1]|-{\textstyle{\circlearrowleft}}}
\newcommand{\ard}[1]{\ar@{.>}[#1]}
\newcommand{\mt}[1]{\ar@{|->}[#1]}
\newcommand{\inm}[1]{\ar@{}[#1]|-{\in}}
\newcommand{\inr}{\ar@{}[d]|-{\rotatebox[origin=c]{-90}{$\in$}}}
\newcommand{\inl}{\ar@{}[u]|-{\rotatebox[origin=c]{90}{$\in$}}}

\newcommand{\sumr}[2]{\sum_{\scriptsize \begin{array}{c}{#1}\\{#2}\end{array}}}%sum with 2 rows
\newcommand{\sumo}[2]{\sum_{#1=1}^{#2}}
\newcommand{\sumz}[2]{\sum_{#1=0}^{#2}}

\newcommand{\beq}[1]{\begin{equation}\llabel{#1}}
\newcommand{\eeq}[0]{\end{equation}}
\newcommand{\bal}[0]{\begin{align*}}
\newcommand{\eal}[0]{\end{align*}}%this doesn't work; i don't know why
\newcommand{\ban}[0]{\begin{align}}
\newcommand{\ean}[0]{\end{align}}

\newcommand{\fixme}[1]{{\color{red}#1}}
\newcommand{\llabel}[1]{\label{#1}\text{\fixme{\tiny#1}}}

\newcommand{\arxiv}[1]{\url{http://www.arxiv.org/abs/#1}}

\allowdisplaybreaks[2]

\DeclareFontFamily{U}{wncy}{}
    \DeclareFontShape{U}{wncy}{m}{n}{<->wncyr10}{}
    \DeclareSymbolFont{mcy}{U}{wncy}{m}{n}
    \DeclareMathSymbol{\Sh}{\mathord}{mcy}{"58}

\title{Spectral Filtering for  General Linear Dynamical Systems}

\date{}
\author{
  Elad Hazan$^1$ \qquad Holden Lee$^2$ \qquad Karan Singh$^1$ \qquad Cyril Zhang$^1$ \qquad Yi Zhang$^1$ \\
  \\
  $^1$ Department of Computer Science\\
  $^2$ Department of Mathematics\\
  Princeton University \\
  \texttt{\{ehazan,holdenl,karans,cyril.zhang,yz7\}@princeton.edu}
}

\begin{document}
\maketitle

\begin{abstract}
We give a polynomial-time algorithm for learning latent-state linear dynamical systems without system identification, and without assumptions on the spectral radius of the system's transition matrix. The algorithm extends the recently introduced technique of spectral filtering, previously applied only to systems with a symmetric transition matrix, using a novel convex relaxation to allow for the efficient identification of phases.
\end{abstract}

\section{Introduction}

Linear dynamical systems (LDSs) are a cornerstone of signal processing and time series analysis. The problem of predicting the response signal arising from a LDS is a fundamental problem in machine learning, with a history of more than half a century.

% \begin{figure*}[ht]
% \begin{center}
% \centerline{\includegraphics[scale=0.2]{LDS.pdf}}
% \caption{Linear Dynamical System with hidden state.}
% \label{ldsdef}
% \end{center}
% \end{figure*}

An LDS is given by matrices $(A,B,C,D)$. Given a sequence of inputs $\{x_t\}$, the output $\{y_t\}$ of the system is governed by the linear equations
\begin{align*}
h_t &= Ah_{t-1} + Bx_t + \eta_t \\
y_t &= Ch_t + Dx_t + \xi_t,
\end{align*}
where $\eta_t, \xi_t$ are noise vectors, and $h_t$ is a hidden (latent) state. It has been observed numerous times in the literature that if there is no hidden state, or if the transition matrices are known, then the formulation is essentially convex and amenable to efficient optimization. 

In this paper we are concerned with the general and more challenging case, arguably the one which is more applicable as well, in which the hidden state is not observed, and the system dynamics are unknown to the learner. In this setting, despite the vast literature on the subject from various communities, there is a lack of provably efficient methods for learning the LDS without strong generative or other assumptions.

Building on the recent spectral filtering technique from \cite{HKZ17}, we develop a more general convex relaxation theorem for LDSs, resulting in an efficient algorithm for the LDS prediction problem in the general setting. Our algorithm makes online predictions which are close (in terms of mean squared error) to those of the optimal LDS in hindsight.

\subsection{Our contributions}

Let $x_1,\ldots,x_T \in \reals^n$ and $y_1,\ldots,y_T \in \reals^{m}$ be the inputs and outputs of a linear dynamical system. Our main result is a polynomial-time algorithm that predicts $\yhat_t$ given all previous input and feedback $(x_{1:t}, y_{1:t-1})$, and competes with the predictions of the best LDS in terms of mean squared error.
Defining regret as the additional least-squares loss incurred compared to the best predictions $y^*_1, \ldots, y^*_T$:
$$ \mathsf{Regret}(T) := \sum_{t=1}^T \norm{ \yhat_t - y_t }^2  -   \sum_{t=1}^T \norm{ y_t^* - y_t }^2, $$
our main algorithm achieves a bound of
\[ \mathsf{Regret}(T) \leq \tilde{O}(\sqrt{T} ) + K \cdot L. \]

Here, $L$ denotes the inevitable loss incurred by perturbations to the system which cannot be anticipated by the learner, which are allowed to be adversarial. The constant in the $\tilde O(\cdot)$, as well as $K$, depend only polynomially on the dimensionality of the system, the norms of the inputs and outputs, and certain natural quantities related to the transition matrix $A$. Additionally, the running time of our algorithm is polynomial in all natural parameters of the problem. 

In comparison to previous approaches, we note:
\begin{itemize}
\item
The main feature is that the regret does not depend on the spectral radius $\rho(A)$ of the system's hidden-state transition matrix. If one allows a dependence on the condition number, then simple linear regression-based algorithms are known to obtain the same result, with time and sample complexity polynomial in $\frac{1}{1-\rho(A)}$.
For a reference, see section 6 of \cite{hardt2016gradient}.

\item
Our algorithm is the first polynomial-time algorithm with this guarantee. However, inefficient algorithms that attain similar regret bounds are also possible, based on the continuous multiplicative-weights algorithm (see \cite{CBL}, as well as the EWOO algorithm in \cite{HAK}). These methods, mentioned briefly in \cite{HKZ17}, basically amount to discretizing the entire parameter space of LDSs, and take time exponential in the system dimensions.
\end{itemize}

\subsection{Techniques} 

The approach of \cite{HKZ17} makes use of the eigendecomposition of the symmetric transition matrix $A$: they show that impulse response function for a real scalar is close to a fixed low-rank subspace defined by the top eigenvectors of a certain Hankel matrix. For the case of a general (asymmetric) LDS, however, the eigenvalues of $A$ are complex-valued, and the subspace property no longer holds. 

The primary difficulty is that the complex-valued Hankel matrix corresponding to complex power does not have the same approximate low-rank property. Instead, we use the fact that the approximation result with real eigenvalues holds if all of the phases of the complex eigenvalues have been identified and taken into account. However, identifying the phases and simultaneously learning the other relevant parameters of the predictor is a non-convex learning problem. We give a novel convex relaxation of this simultaneous learning problem that may be useful in other regret minimization settings.

A second important component is incorporating an autoregressive model, i.e. predicting the current output as a learned linear function of the past $\tau$ inputs and outputs. Intuitively, for a general LDS, regressing only on the previous output $y_{t-1}$ (as was done in \cite{HKZ17} for symmetric LDSs) is problematic, because no single linear map can cancel out the effect of multiple complex rotations. Instead, we use an autoregressive time window of $d$, the dimension of the hidden state.
The autoregressive component accomplishes two objectives: first, it makes the system responsive to noise on the hidden state, and second, it ensures that the additional dependence of $y_t$ on the (in general long) history of inputs $x_1,\ldots, x_t$ is bounded, and hence the wave-filtering method will succeed.

%HL: Added AR <> LDS
Autoregressive models have been studied and used extensively; however, we are not aware of formal analyses on how well they can approximate a general LDS with a hidden state. As part of our analysis, we show that the error or ill-conditionedness of this approximation can be bounded in terms of norms of certain natural polynomials related to the dynamical system.

\subsection{Related work}

The prediction problems of time series for linear dynamical systems was defined in the seminal work of Kalman \cite{kalman1960new}, who introduced the Kalman filter as a recursive least-squares solution for maximum likelihood estimation (MLE) of Gaussian perturbations to the system. For more background see the classic survey \cite{ljung1998system}, and the extensive overview of recent literature in \cite{hardt2016gradient}. 

For a linear dynamical system with no hidden state, the system is identifiable by a convex program and thus well understood (see \cite{dean2017sample,abbasi2011regret}, who  address sample complexity issues and regret for system identification and linear-quadratic control in this setting).

Various approaches have been proposed to learn the system in the case that the system is unknown, however most  
Ghahramani and Roweis \cite{roweis1999unifying} suggest using the EM algorithm to learn the parameters of an LDS. This approach remains widely used, but is inherently non-convex and can get stuck in local minima. 
Recently \cite{hardt2016gradient} show that for a restricted class of systems, gradient descent (also widely used in practice, perhaps better known in this setting as backpropagation) guarantees polynomial convergence rates and sample complexity in the batch setting. Their result applies essentially only to the SISO case, depends polynomially on the spectral gap, and requires the signal to be generated by an LDS.

In recent work, \cite{HKZ17} show how to efficiently learn an LDS in the online prediction setting, without any generative assumptions, and without dependence on the condition number. Their new methodology, however, was restricted to LDSs with symmetric transition matrices. For the structural result, we use the same results from the spectral theory of Hankel matrices; see \cite{beckermann2017singular,hilbert1894beitrag,choi1983tricks}.
As discussed in the previous section, obtaining provably efficient algorithms for the general case is significantly more challenging.

We make use of linear filtering, or linear regression on the past observations as well as inputs, as a subroutine for future prediction. This technique is well-established in the context of autoregressive models for time-series prediction that have been extensively studied in the learning and signal-processing literature, see e.g. \cite{Hamilton94,BoxJenRe94,BroDav09,kuznetsov16,DBLP:conf/colt/AnavaHMS13,tzachi}.

The recent success of recurrent neural networks (RNNs) for tasks such as speech and language modeling has inspired a resurgence of interest in linear dynamical systems \cite{hardt2016gradient,belanger2015linear}.

\section{Preliminaries}
\subsection{Setting}
\label{sec:setting}

A \emph{linear dynamical system} $\Theta = (A,B,C,D)$, with initial hidden state $h_0 \in \reals^d$, specifies a map from \emph{inputs} $x_1, \ldots, x_T \in \reals^n$ to \emph{outputs} (responses) $y_1, \ldots, y_T \in \reals^m$, given by the recursive equations
\begin{align}\label{eq:lds-noise1}
h_t &= Ah_{t-1} + Bx_t + \eta_t \\
y_t &= Ch_t + Dx_t + \xi_t,
\label{eq:lds-noise2}
\end{align}
where $A,B,C,D$ are matrices of appropriate dimension, and $\eta_t, \xi_t$ are noise vectors.

We make the following assumptions to characterize the ``size'' of an LDS we are competing against:
\begin{enumerate}
\item Inputs and outputs and bounded: $\norm{x_t}_2 \leq R_x, \norm{y_t}_2 \leq R_y$.\footnote{Note that no bound on $\norm{y_t}$ is required for the approximation theorem; $R_y$ only appears in the regret bound.}
\item The system is Lyapunov stable, i.e., the largest singular value of $A$ is at most 1: $\rh(A) \leq 1$. Note that we do not need this parameter to be bounded away from $1$.
%HL: Eigenvalue, not singular value! They are not the same.
\item $A$ is diagonalizable by a matrix with small entries: $A = \Psi \Lambda \Psi^{-1}$, with $\ve{\Psi}_F\ve{\Psi^{-1}}_F \le R_\Psi$. 
%$\max( \norm{\Psi}_F, \norm{\Psi^{-1}}_F ) \leq R_\Psi$.
Intuitively, this holds if the eigenvectors corresponding to larger eigenvalues aren't close to linearly dependent.
\item $B,C,D$ have bounded spectral norms: $\ve{B}_2,\ve{C}_2,\ve{D}_2 \le R_\Te$.
\item Let $S=\set{\fc{\al}{|\al|}}{\al\text{ is an eigenvalue of }A}$ be the set of phases of all eigenvalues of $A$. There exists a monic polynomial $p(x)$ of degree $\tau$ such that $p(\om)=0$ for all $\om\in S$, the $L^1$ norm of its coefficients is at most $R_1$, and the $L^\iy$ norm is at most $R_\iy$. We will explain this condition in Section~\ref{sec:approx}.
\end{enumerate}

In our regret model, the adversary chooses an LDS $(A,B,C,D)$, and has a budget $L$.
The dynamical system produces outputs given by the above equations, where the noise vectors $\eta_t, \xi_t$ are chosen adversarially, subject to a budget constraint: $\sum_{t=1}^T \norm{\eta_t}^2 + \norm{\xi_t}^2 \leq L$.

Then, the online prediction setting is identical to that proposed in \cite{HKZ17}. For each iteration $t = 1, \ldots, T$, the input $x_t$ is revealed, and the learner must predict a response $\hat y_t$. Then, the true $y_t$ is revealed, and the learner suffers a \emph{least-squares} loss of $\norm{y_t - \hat y_t}^2$. Of course, if $L$ scales with the time horizon $T$, it is information-theoretically impossible for an online algorithm to incur a loss sublinear in $T$, even under non-adversarial (e.g. Gaussian) perturbations. Thus, our end-to-end goal is to track the LDS with loss that scales with the total magnitude of the perturbations, independently of $T$.

%\hl{Mention connection to $H^\iy$, robust learning of LDS, min-max formulation.}
This formulation is fundamentally a min-max problem: given a limited budget of perturbations, an adversary tries to maximize the error of the algorithm's predictions, while the algorithm seeks to be robust against any such adversary. This corresponds to the $H^\iy$ notion of robustness in the control theory literature; see Section 15.5 of \cite{zhou1996robust}.

\subsection{The spectral filtering method}
The \emph{spectral filtering} technique is introduced in \cite{HKZ17}, which considers a spectral decomposition of the derivative of the impulse response function of an LDS with a symmetric transition matrix. We derive Theorem~\ref{thm:arma-wf}, an analogue of this result for the asymmetric case, using some of the technical lemmas from this work.

A crucial object of consideration from \cite{HKZ17} is the set of \emph{wave-filters} $\phi_1, \ldots, \phi_k$, which are the top $k$ eigenvectors of the deterministic Hankel matrix $Z_T \in \reals^{T\times T}$, whose entries are given by $Z(i,j) = \frac{2}{(i+j)^3 - (i+j)}$. Bounds on the $\eps$-rank of positive semidefinite Hankel matrices can be found in \cite{beckermann2017singular}. Like Algorithm~1 from \cite{HKZ17}, the algorithm will ``compress'' the input time series using a time-domain convolution of the input time series with filters derived from these eigenvectors.

\subsection{Notation for matrix norms}
We will consider a few ``mixed'' $\ell_p$ matrix norms of a 4-tensor $M$,
whose elements are indexed by $M(p,h,j,i)$ (the roles and bounds of these indices will be introduced later). For conciseness, whenever the norm of such a 4-tensor is taken, we establish the notation for the mixed matrix norm
\[ \norm{M}_{2,q} := \left[ \sum_p \left( \sum_{h,i,j} M(p,h,j,i)^2 \right)^{q/2} \right]^{1/q}, \]
and the limiting case
\[ \norm{M}_{2,\infty} := \max_p \sqrt{ \sum_{h,i,j} M(p,h,j,i)^2 }. \]
These are the straightforward analogues of the matrix norms defined in \cite{kakade2012regularization}, and appear in the regularization of the online prediction algorithm.

\section{Algorithm}
To define the algorithm, we specify a reparameterization of linear dynamical systems.
To this end, we define a \emph{pseudo-LDS}, which pairs a subspace-restricted linear model of the impulse response with an autoregressive model:

\begin{definition}\label{eq:pseudo-LDS}
A \emph{pseudo-LDS} $\hat\Theta = (M,N,\beta,P)$ is given by two 4-tensors $M,N \in \reals^{W \times k \times n \times m}$
a vector $\beta \in \reals^\tau$, and matrices $P_0,\ldots, P_{\tau -1}\in \R^{m\times n}$.
Let the \emph{prediction} made by $\hat\Theta$, which depends on the entire history of inputs $x_{1:t}$ and $\tau$ past outputs $y_{t-1:t-\tau}$ be given by

\ifdefined\VIX
    \begin{align}
     \notag &y(\hat\Theta; x_{1:t}, y_{t-1:t-\tau})(:) 
     :=
      \sum_{u=1}^\tau \beta_u y_{t-u} +
    \sumz j{\tau-1} P_j x_{t-j}\\
     \notag &\quad +
    \sum_{p=0}^{W-1} 
    \sum_{i=1}^n \sum_{h=1}^k
    \sum_{u=1}^T 
    \Bigg[ 
    \Bigg(
    M(p,h,:,i) \cos\pf{2\pi u p}{W} 
    +N(p,h,:,i) \sin\pf{2\pi u p}{W}\Bigg)
    \si_h^{\rc 4}
    \phi_h(u) x_{t-u}(i)\Bigg] 
    \end{align}
\else
   \begin{align}
     \notag &y(\hat\Theta; x_{1:t}, y_{t-1:t-\tau})(:) 
     :=
      \sum_{u=1}^\tau \beta_u y_{t-u} +
    \sumz j{\tau-1} P_j x_{t-j}\\ 
     \notag &+
    \sum_{p=0}^{W-1} 
    \sum_{i=1}^n \sum_{h=1}^k
    \sum_{u=1}^T 
    \Bigg[ 
    \Bigg(
    M(p,h,:,i) \cos\pf{2\pi u p}{W}\\ 
    &+N(p,h,:,i) \sin\pf{2\pi u p}{W}\Bigg)
    \si_h^{\rc 4}
    \phi_h(u) x_{t-u}(i)\Bigg] \label{eq:pseudo-LDS}
    \end{align}
\fi

\end{definition}
Here, $\phi_1, \ldots, \phi_k \in \reals^T$ are the top $k$ eigenvectors, with eigenvalues $\si_1, \ldots, \si_k$, of $Z_T$. These can be computed using specialized methods \cite{boley1998fast}.
Some of the dimensions of these tensors are parameters to the algorithm, which we list here:
\begin{itemize}
\item Number of filters $k$.
\item Phase discretization parameter $W$.
\item Autoregressive parameter $\tau$.
\end{itemize}
Additionally, we define the following:
\begin{itemize}
\item Regularizer $R(M,N,\beta,P) := \norm{M}_{2,q}^2 + \norm{N}_{2,q}^2 + \norm{\beta}_{q'}^2 + \sum_{j=1}^\tau \norm{P_j}_F^2$, where $q = \frac{\ln(W)}{\ln(W)-1}$, and $q' = \frac{\ln(\tau)}{\ln(\tau)-1}$.
\item Composite norm $\norm{ (M,N,\beta,P) } := \norm{M}_{2,1} + \norm{N}_{2,1} + \norm{\beta}_1 + \sqrt{ \sum_{j=1}^\tau \norm{P_j}_F^2 }$.
\item Composite norm constraint $R_{\hat \Theta}$, and
the corresponding set of pseudo-LDSs $\mathcal{K} = \{\hat\Theta : \norm{\hat\Theta} \leq R_{\hat\Theta} \}$.
\end{itemize}

Crucially, $y(\hat\Theta; x_{1:t}, y_{t-1:t-d})$ is linear in each of $M,N,P,\beta$; consequently, the least-squares loss $\norm{y(M;x_{1:t}) - y}^2$
is convex, and can be minimized in polynomial time.
To this end, our online prediction algorithm
is follow-the-regularized-leader (FTRL), which requires the solution of a convex program at each iteration. We choose this regularization to obtain the strongest theoretical guarantee, and provide a brief note in Section~5 on alternatives to address performance issues.

\begin{figure}[ht]
\vskip 0.2in
\begin{center}
\ifdefined\VIX
    \centerline{\includegraphics[width=0.6\linewidth]{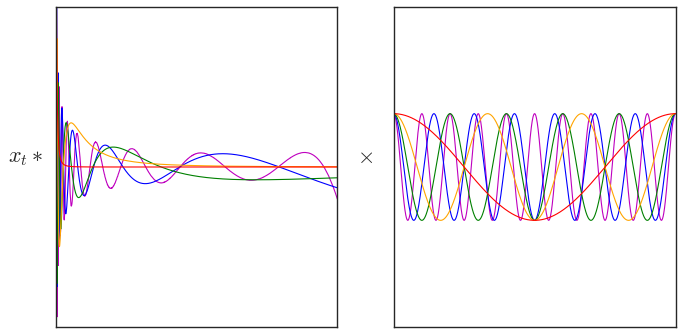}}
\else
   \centerline{\includegraphics[width=\linewidth]{filters.png}}
\fi
\caption{Schematic of the ``compression'' used by a pseudo-LDS to approximate an LDS; compare with Figure 1 in \cite{HKZ17}. Each block of the 4-tensor $M(p,h,:,:)$ (resp. $N(p,h,:,:)$) specifies the linear dependence of $y_t$ on a convolution by $\phi_h$ (left) times the sinusoid $\cos(2\pi up/W)$ (right; resp. $\sin(\cdot)$).}
\label{fig:filters}
\end{center}
\vskip -0.2in
\end{figure}

At a high level, our algorithm works by first approximating the response of an LDS by an autoregressive model of order $(\tau,\tau)$, then refining the approximation using wave-filters with a \emph{phase} component. Specifically, the blocks of $M$ and $N$ corresponding to filter index $h$ and phase index $p$ specify the linear dependence of $y_t$ on a certain convolution of the input time series, whose kernel is the pointwise product of $\phi_h$ and a sinusoid with period $W/p$. The structural result which drives the theorem is that the dynamics of any true LDS are approximated by such a pseudo-LDS, with reasonably small parameters and coefficients.

Note that the autoregressive component in our definition of a pseudo-LDS is slightly more restricted than multivariate autoregressive models: the coefficients $\beta_j$ are \emph{scalar}, rather than allowed to be arbitrary matrices. These options are interchangeable for our purposes, without affecting the asymptotic regret; we choose to use scalar coefficients for a more streamlined analysis.

The online prediction algorithm is fully specified in Algorithm~\ref{alg:asym-lds}; the parameter choices that give the best asymptotic theoretical guarantees are specified in the appendix, while typical realistic settings are outlined in Section~5.

\begin{algorithm}
\caption{Online wave-filtered regression with phases}
\label{alg:asym-lds}
\begin{algorithmic}[1]
\STATE Input: time horizon $T$, parameters $k,W,\tau,R_{\hat \Theta}$, regularization weight $\eta$.
\STATE Compute $\{(\sigma_j, \phi_j)\}_{j=1}^k$, the top $k$ eigenpairs of $Z_T$.
\STATE Initialize $\hat\Theta_1 \in \mathcal{K}$ arbitrarily.
\FOR{$t = 1, \ldots, T$}
\STATE Predict $\hat y_t := y(\hat\Theta_t;x_{1:t};y_{t-1:t-\tau})$.
\STATE Observe $y_t$. Suffer loss $\norm{y_t - \hat y_t}^2$.
\STATE Solve FTRL convex program:
\ifdefined\VIX
  \begin{align*}
  \hat\Theta_{t+1} \leftarrow \argmin_{\hat\Theta \in \mathcal{K}} \sum_{u=0}^{t-1} \norm{ y(\hat\Theta; x_{1:u}, y_{u-1:u-\tau} ) - y_u }^2 + \frac{1}{\eta}R(\hat\Theta).
  \end{align*}
\else
  \begin{align*}
  \hat\Theta_{t+1} \leftarrow \argmin_{\hat\Theta \in \mathcal{K}} \sum_{u=0}^{t-1} \norm{ y(\hat\Theta; x_{1:u}, y_{u-1:u-\tau} ) - y_u }^2 \\ + \frac{1}{\eta}R(\hat\Theta).
  \end{align*}
\fi
\ENDFOR
\end{algorithmic}
\end{algorithm}

\section{Analysis}
\begin{theorem}[Main; informal]\label{thm:main}
Consider a LDS with noise (given by~\eqref{eq:lds-noise1} and~\eqref{eq:lds-noise2}) satisfying the assumptions in Section~\ref{sec:setting}, where total noise is bounded by $L$. 
Then there is a choice of parameters such that
%with parameters $k=O\pa{\poly\log\pa{T, R_\Te, R_\Psi, R_1, R_x, \rc{\ep}}}$, $P=O\pa{\poly(\tau, R_\Te, R_\Psi, R_1, R_x, T)}$, 
%\hl{what is the regularization weight?} 
Algorithm~\ref{alg:asym-lds} learns a pseudo-LDS $\hat \Te$ whose predictions $\hat y_t$ satisfy
\begin{align}\label{eq:main}
\sumo{t}T \ve{\hat y_t - y_t}^2 &\le 
%O\pa{\poly(R_1, R_x, R_y, R_\Theta, R_\Psi, m, n, d, \log T) \sqrt{T} + R_\iy^2 \tau^{3} R_\Te^2 R_\Psi^2 L}
\tilde{O}\pa{\poly(R, d') \sqrt{T} + R_\iy^2 \tau^{3} R_\Te^2 R_\Psi^2 L}
\end{align} where $R_1, R_x, R_y, R_\Theta, R_\Psi\leq R$, $m, n, d\leq d'$.
\end{theorem}
There are three parts to the analysis, which we outline in the following subsections: proving the approximability of an LDS by a pseudo-LDS, bounding the regret incurred by the algorithm against the best pseudo-LDS, and finally analyzing the effect of noise $L$. The full proofs are in Appendices~\ref{app:approx},~\ref{section:appendix-regret-proof}, and~\ref{app:main}, respectively.

\subsection{Approximation theorem for general LDSs}
%\begin{theorem}[Approximate convex relaxation; informal]
%\end{theorem}
We develop a more general analogue of the structural result from \cite{HKZ17}, which holds for 
systems with asymmetric transition matrix $A$.
%systems with a symmetric transition matrix $A$.
\label{sec:approx}

\begin{theorem}[Approximation theorem; informal]\label{thm:arma-wf}
Consider an noiseless LDS (given by~\eqref{eq:lds-noise1} and~\eqref{eq:lds-noise2} with $\eta_t,\xi_t=0$) satisfying the assumptions in Section~\ref{sec:setting}. 
%Let $S=\set{\fc{\al}{|\al|}}{\al\text{ is an eigenvalue of }A}$ be the set of phases of all eigenvalues of $A$. Further assume there exists a monic polynomial $p(x)$ of degree $\tau$ such that $p(\om)=0$ for all $\om\in S$, and let $R_1$ be the $L^1$ norm of its coefficients. 

There is $k=O\pa{\poly\log\pa{T, R_\Te, R_\Psi, R_1, R_x, \rc{\ep}}}$, $W=O\pa{\poly(\tau, R_\Te, R_\Psi, R_1, R_x, T)}$ and a pseudo-LDS $\Te$ of norm $O(\poly(R_\Te, R_\Psi, R_1, \tau, k))$ such that 
%the pseudo-LDS $(M,N,\be,P)$ 
$\Te$ approximates $y_t$ to within $\ep$ for $1\le t\le T$:
\begin{align}%M,N,\be,P_{0:\tau-1}
\ve{y(\Te;x_{1:t},y_{t-1:t-\tau}) - y_t}&\le \ep.
\end{align}
\end{theorem}
For the formal statement (with precise bounds) and proof, see Appendix~\ref{app:ar-wave}.
In this section we give some intuition for the conditions and an outline of the proof.

First, we explain the condition on the polynomial $p$. As we show in Appendix~\ref{sec:ar} we can predict using a pure autoregressive model, without wavefilters, if we require $p$ to have all eigenvalues of $A$ as roots (i.e., it is divisible by the minimal polynomial of $A$). However, the coefficients of this polynomial could be very large. The size of these coefficients will appear in the bound for the main theorem, as using large coefficients in the predictor will make it sensitive to noise.

Requiring $p$ only to have the phases of eigenvalues of $A$ as roots can decrease the coefficients significantly. As an example, consider if $A$ has many $d/3$ distinct eigenvalues with phase $1$, and similarly for $\om$, and $\ol{\om}$, and suppose their absolute values are close to 1. Then the minimal polynomial is approximately $\prod (x-1)^{\fc d3} (x-\om)^{\fc d3} (x-\ol{\om})^{\fc d3}$ which can have coefficients as large as $\exp(\Om(d))$. On the other hand, for the theorem we can take $p(x)=(x-1)(x-\om)(x-\ol{\om})$ which has degree 3 and coefficients bounded by a constant. Intuitively, the wavefilters help if there are few distinct phases, or they are well-separated (consider that if the phases were exactly the $d$th roots of unity, that $p$ can be taken to be $x^d-1$). Note that when the roots are real, we can take $p=x-1$ and the analysis reduces to that of~\cite{HKZ17}.

We now sketch a proof of Theorem~\ref{thm:arma-wf}. Motivation is given by the Cayley-Hamilton Theorem, which says that if $p$ is the characteristic polynomial of $A$, then $p(A)=O$. This fact tells us that the $h_t=A^th_0$ satisfies a linear recurrence of order $\tau=\deg p$: if $p(x) = x^\tau + \sumo t{\tau} \be^j x^{\tau-j}$, then $h_t +\sumo t{\tau} \be_j h_{t-j}=0$. 

If $p$ has only the phases as the roots, then $h_t +\sumo t{\tau} \be_j h_{t-j}\ne 0$ but can be written in terms of the wavefilters.
%Given a sequence $h_t=A^t h_0$, 
Consider for simplicity the 1-dimensional (complex) LDS $y_t = \al y_{t-1}+x_t$, and let $\al = r\om$ with $|\om|=1$. Suppose $p(x) = x^\tau + \sumo t{\tau} \be^j x^{\tau-j}=0$ and $p(\om)=0$. In general the LDS is a ``sum'' of LDS's that are in this form. 
Summing the past $\tau$ terms with coefficients given by $\be$,

\begin{footnotesize}
\begin{center}
\begin{tabular}{rp{1mm}p{10mm}ccp{5mm}}
$y_{t}=$ & $x_{t}$ & $+\alpha x_{t-1}$ & $+\cdots$ & $+\alpha^{\tau}x_{t-\tau}$ & $+\cdots$\tabularnewline
+$\beta_{1}(y_{t-1}=$ &  & $x_{t-1}$ & $+\cdots$ & $+\alpha^{\tau-1}x_{t-\tau}$ & $+\cdots)$\tabularnewline
$\vdots$ &  &  & $\vdots$ & $\vdots$ & \tabularnewline
$+\beta_{\tau}(y_{t-\tau}=$ &  &  &  & $x_{t-\tau}$ & $+\cdots)$\tabularnewline
\end{tabular}
\end{center}
\end{footnotesize}

The terms $x_t,\ldots, x_{t-\tau+1}$ can be taken care of by linear regression. Consider a term $x_j$, $j<t-\tau$ in this sum. The coefficient is $\al^{j-(t-\tau)}(\al^\tau + \be_1\al^{\tau-1}+\cdots + \be_\tau) $. Because $p(\om)=0$, this can be written as 
\begin{align}
\al^{j-(t-\tau)} ((\al^\tau - \om^\tau) + \be_1(\al^{\tau-1}-\om^{\tau-1})+\cdots).
\end{align}
Factoring out $1-r$ from each of these terms show that $y_t + \be_1 y_{t-1} + \cdots + \be_\tau y_{t-\tau}$ can be expressed as a function of a convolution of the vector $((1-r)r^{t-1}\om^{t-1})$ with $x$. The wavefilters were designed precisely to approximate the vector $\mu(r) = ((1-r)r^{t-1})_{1\le t\le T}$ well, hence $y_t + \be_1 y_{t-1} + \cdots + \be_\tau y_{t-\tau}$ %the terms for $j\le t-\tau$ 
can be approximated using the wavefilters multiplied by phase and convolved with $x$. Note that the $1-r$ is necessary in order to make the $L^2$ norm of $((1-r)r^{t-1})_{1\le t\le T}$ bounded, and hence ensure the wavefilters have bounded coefficients.

\subsection{Regret bound for pseudo-LDSs}
As an intermediate step toward the main theorem, we show a regret bound on the total least-squares prediction error made by Algorithm~\ref{alg:asym-lds}, compared to the best \emph{pseudo-LDS} in hindsight.

\begin{theorem}[FTRL regret bound; informal]
\label{thm:ftrl}
Let $y^*_1, \ldots, y^*_T$ denote the predictions made by the fixed pseudo-LDS minimizing the total squared-norm error. Then, there is a choice of parameters for which the decision set $\mathcal{K}$ contains all LDSs which obey the assumptions from Section 2.1,
for which the predictions $\hat y_1, \ldots, \hat y_T$ made by Algorithm~\ref{alg:asym-lds} satisfy
\[ \sum_{t=1}^T \norm{\hat y_t - y_t}^2 - \sum_{t=1}^T \norm{\hat y_t^* - y_t}^2 \leq
\tilde{O}\pa{\poly(R, d') \sqrt{T}}. \]
where $R_1, R_x, R_y, R_\Theta, R_\Psi\leq R$, $m,n,d\leq d'$.
\end{theorem}

The regret bound follows by applying the standard regret bound of follow-the-regularized-leader (see, e.g. \cite{OCObook}). However, special care must be taken to ensure that the gradient and diameter factors incur only a $\poly\log(T)$ factor, noting that the discretization parameter $W$ (one of the dimensions of $M$ and $N$) must depend polynomially on $T/\eps$ in order for the class of pseudo-LDSs to approximate true LDSs up to error $\eps$. To this end, we use a modification of the strongly convex matrix regularizer found in \cite{kakade2012regularization}, resulting in a regret bound with logarithmic dependence on $W$.

Intuitively, this is possible due to the $d$-sparsity (and thus $\ell_1$ boundedness) of the phases of true LDSs, which transfers to an $\ell_1$ bound (in the phase dimension only) on pseudo-LDSs that compete with LDSs of the same size. This allows us to formulate a second convex relaxation, on top of that of wave-filtering, for simultaneous identification of eigenvalue phase and magnitude. For the complete theorem statement and proof, see Appendix~\ref{section:appendix-regret-proof}.

We note that the regret analysis can be used directly with the approximation result for autoregressive models (Theorem~\ref{thm:arma}), without wave-filtering. This way, one can straightforwardly obtain a sublinear regret bound against autoregressive models with bounded coefficients. However, for the reasons discussed in Section~\ref{sec:approx}, the wave-filtering technique affords us a much stronger end-to-end result.

\subsection{Pseudo-LDSs compete with true LDSs}
%To prove the main theorem, we first need to analyze the approximation when there is noise. Compared to the noiseless case, as analyzed in Theorem~\ref{thm:arma-wf-precise}, we incur an additional term equal to the size of the perturbation times a {\it competitive ratio} depending on the dynamical system.

Theorem~\ref{thm:arma-wf} shows that there exists a pseudo-LDS approximating the actual LDS to within $\ep$ in the noiseless case. We next need to analyze the
best approximation when there is noise. 
We show in Appendix~\ref{app:main} (Lemma~\ref{lem:perturb}) that if the noise is bounded ($\sumo tT \ve{\eta_t}_2^2 + \ve{\xi_t}_2^2\le L$), we incur an additional term equal to the size of the perturbation $\sqrt L$ times a {\it competitive ratio} depending on the dynamical system, for a total of $R_\iy \tau^{\fc 32} R_\Te R_\Psi \sqrt L$. We show this by showing that any noise has a bounded effect on the predictions of the pseudo-LDS.\footnote{In other words, the prediction error of the pseudo-LDS is stable to noise, and we bound its $H^\iy$ norm.}

Letting $y_t^*$ be the predictions of the best pseudo-LDS, we have
\ifdefined\VIX
  \begin{align}
  \sumo tT \ve{\hat y_{t} - y_{t}}_2^2
  =& \pa{
  \sumo tT\ve{\hat y_{t} - y_{t}}_2^2 - 
  \sumo tT\ve{\hat y_{t}^* - \hat y_{t}}_2^2 
  }+ \sumo tT\ve{\hat y_{t}^* - \hat y_{t}}_2^2.
  \end{align}
\else
	\begin{align}
  \sumo tT \ve{\hat y_{t} - y_{t}}_2^2
  =& \pa{
  \sumo tT\ve{\hat y_{t} - y_{t}}_2^2 - 
  \sumo tT\ve{\hat y_{t}^* - \hat y_{t}}_2^2 
  }\\
  &+ \sumo tT\ve{\hat y_{t}^* - \hat y_{t}}_2^2.
\end{align}

\fi
The first term is the regret, bounded by Theorem~\ref{thm:ftrl} and the second term is bounded by the discussion above, giving the bound in the Theorem~\ref{thm:main}.

For the complete proof, see Appendix~\ref{app:main-pf}.

\section{Experiments}
We exhibit two experiments on synthetic time series, which are generated by randomly-generated ill-conditioned LDSs. In both cases, $A \in \reals^{10\times 10}$ is a block-diagonal matrix, whose 2-by-2 blocks are rotation matrices $[\cos\theta \;\; {-\sin\theta} ; \sin\theta \; \cos\theta]$ for phases $\theta$ drawn uniformly at random. This comprises a hard case for direct system identification: long-term time dependences between input and output, and the optimization landscape is non-convex, with many local minima. Here, $B \in \reals^{10\times 10}$ and $C \in \reals^{2\times 10}$ are random matrices of standard i.i.d. Gaussians. In the first experiment, the inputs $x_t$ are i.i.d. spherical Gaussians; in the second, the inputs are Gaussian block impulses.

\begin{figure*}[ht]
\vskip 0.2in
\begin{center}
\ifdefined\VIX
	\centerline{\includegraphics[width=\linewidth]{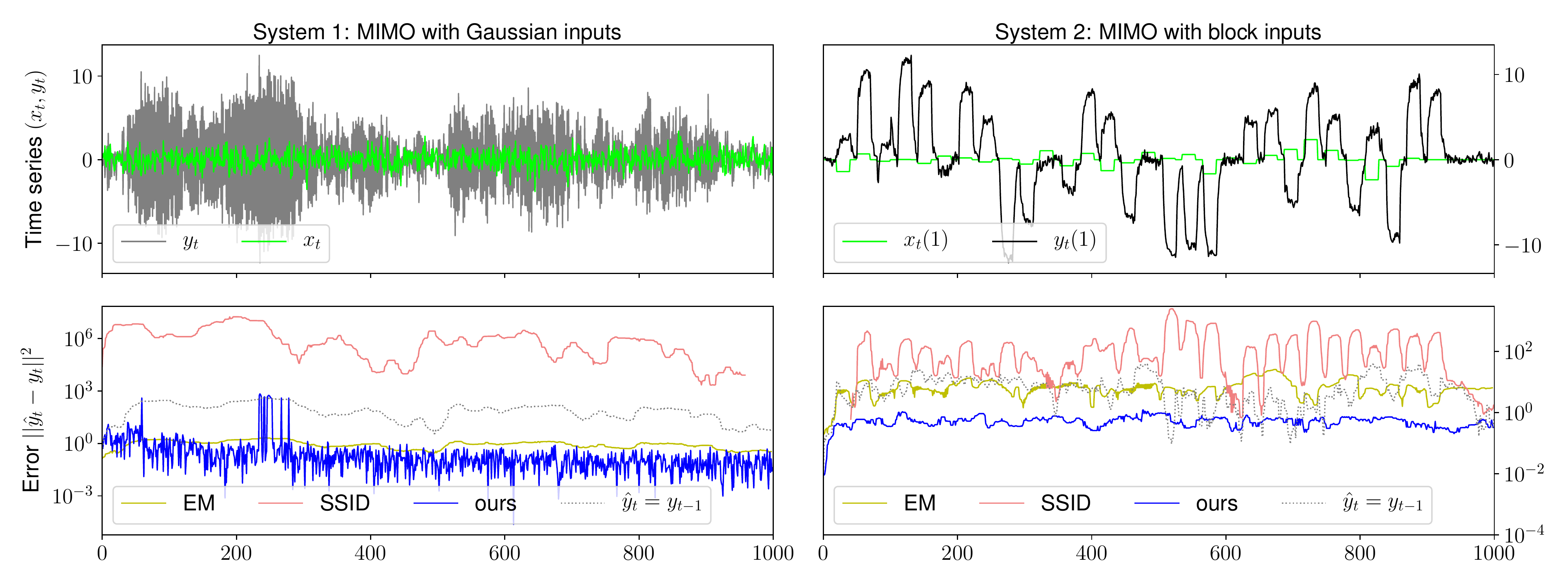}}
\else
	\centerline{\includegraphics[width=\linewidth]{toy.pdf}}
\fi
\caption{Performance of Algorithm~\ref{alg:asym-lds} on synthetic 10-dimensional LDSs.
For clarity, error plots are smoothed by a median filter. \textcolor{blue}{Blue} = ours, \textcolor[rgb]{0.7,0.7,0}{yellow} = EM, \textcolor{red}{red} = SSID, \textbf{black} = true responses, \textcolor{green}{green} = inputs, dotted lines = ``guess the previous output'' baseline. Horizontal axis is time. \emph{Left:} Gaussian inputs; SSID fails to converge, while EM finds a local optimum. \emph{Right:} Block impulse inputs; both baselines find local optima.}
\label{fig:toy-experiments}
\end{center}
\vskip -0.2in
\end{figure*}

We make a few straightforward modifications to Algorithm~\ref{alg:asym-lds}, for practicality. First, we replace the scalar autoregressive parameters with matrices $\beta_j \in \reals^{m \times m}$. Also, for performance reasons, we use ridge regularization instead of the prescribed pseudo-LDS regularizer with composite norm constraint.
We choose an autoregressive parameter of $\tau = d = 10$ (in accordance with the theory), and $W = 100$.

As shown in Figure~2, our algorithm significantly outperforms the baseline methods of system identification followed by Kalman filtering. The EM and subspace identification (SSID; see \cite{van2012subspace}) algorithms finds a local optimum; in the experiment with Gaussian inputs, the latter failed to converge (left).

We note that while the main online algorithm from \cite{HKZ17}, Algorithm~\ref{alg:asym-lds} is significantly faster than baseline methods, ours is not. The reason is that we incur at least an extra factor of $W$ to compute and process the additional convolutions. To remove this phase discretization bottleneck, many heuristics are available for phase identification; see Chapter 6 of \cite{ljung1998system}.

\section{Conclusion}

We gave the first, to the best of our knowledge, polynomial-time algorithm for prediction in the general LDS setting without dependence on the spectral radius parameter of the underlying system.  Our algorithm combines several techniques, namely the recently introduced wave-filtering method, as well as convex relaxation and linear filtering. 

One important future direction is to improve the regret in the setting of (non-adversarial) Gaussian noise. In this setting, if the LDS is explicitly identified, the best predictor is the Kalman filter, which, when unrolled, depends on feedback for {\it all} previous time steps, and only incurs a cost $O(L)$ from noise in~\eqref{eq:main}. It is of great theoretical and practical interest to compete directly with the Kalman filter without system identification.

\bibliographystyle{plain}
\bibliography{ICML18/main}

%----- APPENDIX BELOW

\appendix
\section{Proof of approximation theorem}
\label{app:approx}
\subsection{Warm-up: Simple autoregressive model}
\label{sec:ar}

As a warm-up, we first establish rigorous regret bounds for an autoregressive model, depending on properties of the the minimal polynomial of the linear dynamical system. Next we will see how introducing wavefilters can improve these bounds.

\begin{theorem}\label{thm:arma}
Consider a LDS $\Te=(A,B,C, h_0=0)$, where $A=\Psi \La \Psi^{-1}$ is diagonalizable with spectral radius $\le 1$.
Let $p(x)$ be a monic polynomial of degree $\tau$ such that $p(A)=0$.\footnote{In other words, the minimal polynomial of $A$ divides $p$. For a diagonalizable matrix, the minimal polynomial is the characteristic polynomial except without repeated zeros.}
 Suppose %$\ve{p}_2\le R$, 
 $\ve{p}_1\le R_1$,\footnote{For a polynomial $p(x) = \sumz j{\tau} \be_{j}x^{\tau-j}$, let
\begin{align}
\ve{p}_1 &= \sumz j{\tau} |\be_{j}|
\end{align}} $\ve{B}_2,\ve{C}_2\le R_\Te$, and $\ve{\Psi}_F\ve{\Psi^{-1}}_F\le R_\Psi$. 

Suppose $y_1,\ldots, y_t$ is generated from the LDS with inputs $x_t$. Then there exist $\be\in \R^\tau$ and $P_0,\ldots, P_{\tau-1}\in \R^{m\times n}$ satisfying
\begin{align}\label{eq:arma-cons1}
\ve{\be}_1 &\le R_1%-1
\\
\label{eq:arma-cons2}
\ve{P_j}_F &\le R_1 R_\Te^2 R_\Psi
\end{align}
such that
\begin{align}
y_t & = -\sum_{j=1}^{\tau} \be_j  y_{t-j} + \sum_{j=0}^{\tau-1} P_j x_{t-j}.
\end{align}
\end{theorem}
\begin{proof}
Unfolding the LDS,
\begin{align}
y_t &=\sumz i{t-1} CA^iB x_{t-i} % + CA^t h_0
\\
%y_{t-d+j} &= \sumz i{t-d+j-1}CA^i B x_{t-d+j-i}\\
%&=\sum_{i=d-j}^{t-1} CA^{j-d+1}x_{t-i}
y_{t-j} &= \sumz i{t-j-1} CA^iB x_{t-j-i} % + CA^{t-j} h_0
=\sum_{i=j}^{t-1} CA^{i-j}B x_{t-i}.
\end{align}
Let $p(x) = \sumz j{\tau} \be_{j}x^{\tau-j}$ (with $\be_0=1$).  Then
\begin{align}
\sumz j{\tau} \be_j  y_{t-j} &= \sumr{0\le j\le \tau}{j\le i\le t-1} \be_j  CA^{i-j} Bx_{t-i} \\
&=\sumz i{\tau-1} \sum_{j=0}^i \be_j  CA^{i-j} Bx_{t-i} + \ub{C\sum_{i=\tau}^{t-1} \pa{\sumz j{\tau} \be_j  A^{i-j}}Bx_{t-i}}{=0}\label{eq:char-poly-y}
\end{align}
using the fact that 
$\sum_{j=0}^{\tau} A^{i-j} = A^{i-j-\tau} p(A)=0$. Writing this in the autoregressive format,
\begin{align}
y_t &= -\sumo j{\tau} \be_j  y_{t-j} + C\sumz i{\tau-1} \sum_{j=0}^i \be_j  CA^{i-j} Bx_{t-i}.
\end{align}
We let $N_i = \sum_{j=0}^i \be_j  CA^{i-j}B$ for $0\le i\le \tau-1$ and check that this satisfies the conditions. Note $\ve{MN}_F\le \min\{\ve{M}_2\ve{N}_F,\ve{M}_F\ve{N}_2\}$. By the $L^1$-$L^\iy$ inequality,
\begin{align}
\ve{P_i}_F =\ve{\sumz ji \be_j  CA^{i-j} B}_F
%&\le \pa{\sumz ji \be_j ^2}\pa{\sumz ji \ve{CA^{i-j} B}_F^2}\\
&\le \pa{\sumz ji |\be_j |} \max_{0\le j\le i} \ve{CA^{i-j}B}_F\\
&\le R_1 \ve{C}_2\ve{\Psi}_F \ve{\La}_2\ve{\Psi^{-1}}_F\ve{B}_2\\
&\le R_1 R_\Te^2 R_\Psi.
%\le R^2R_\Te^2.
\end{align}
\end{proof}

\subsection{Autoregressive model with wavefilters: An approximate convex relaxation for asymmetric LDS}
\label{app:ar-wave}

If we add wavefiltered inputs to the regression, the bounds depend not on the minimal polynomial $p$ of $A$, but rather on the minimal polynomial having all the $\om_\ell$ as zeros, where $\om_\ell$ are the phases of zeros of the charateristic polynomial of $A$. For example, if all the roots are real and distinct, then the polynomial is $x-1$ rather than $\prod_{p(\al)=0} (x-\al)$. This can be an exponential improvement. For example, consider the case where all the $\al$ are close to 1. Then the minimal polynomial is close to $(x-1)^d$, which has coefficients as large as $\exp(\Om(d))$. Note the case of real roots reduces to~\cite{HKZ17}.

First, we introduce some notation for convenience. Note that in~\eqref{eq:pseudo-LDS}, $\sumo uT \phi_h(u) x_{t-u}(i) \cos\pf{2\pi u p}{W}$ is a certain convolution. For $a\in \R^T$ we define
\begin{align}
a^{(\om)}:&= (a_j \om^j)_{1\le j\le T}\\
a^{(\cos,\te)} :&= (a_j \cos(j\te))_{1\le j\le T}\\
a^{(\sin,\te)} :&= (a_j \sin(j\te))_{1\le j\le T}.
\end{align}
so that we can write~\eqref{eq:pseudo-LDS} as
\begin{align}\label{eq:pseudo-LDS2}
 &y(M, N,\beta,P; x_{1:t}, y_{t-1:t-d}) := \sum_{u=1}^\tau \beta_u y_{t-u} +
\sumz j{\tau-1} P_j x_{t-j} \\
&\quad + 
\sum_{p=0}^{P-1} 
\sum_{h=1}^k 
\sum_{u=1}^T 
\pa{M(p,h,:,:)
\si_h^{\rc 4}(\phi_h^{(\cos, \fc{2\pi u p}{P})} * x)_{t-\tau}  +
N(p,h,:,:)
\si_h^{\rc 4}
(\phi_j^{(\sin, \fc{2\pi u p}{P})} * x)_{t-\tau}}  
 \end{align}
 
 We now give a more precise statement of Theorem~\ref{thm:arma-wf} and prove it.
 \begin{theorem}\label{thm:arma-wf-precise}
Consider a LDS $\Te=(A,B,C, h_0=0)$, where $A$ is diagonalizable with spectral radius $\le 1$.
Let the eigenvalues of $A$ be $\al_\ell = \omega_\ell r_\ell$ for $1\le \ell\le d$, where $|\omega_\ell|=1$ and $r_\ell\in \R_{\ge 0}$. Let $S$ be the set of $\om_\ell$. %\footnote{\fixme{Can relax this further---each $\om_\ell$ is within some small enough $\ep$ of an element of $S$.}} 
Let $p(x)$ be a monic polynomial of degree $\tau$ such that $p(\om_\ell)=0$ for each $\ell$.
Suppose $\ve{p}_1\le R_1$, 
 $\ve{B}_2,\ve{C}_2\le R_\Te$,
 and the diagonalization $A=\Psi \La \Psi^{-1}$ satisfies $\ve{\Psi}_F\ve{\Psi^{-1}}_F\le R_\Psi$.

Suppose $y_1,\ldots, y_T$ is generated from the LDS with inputs $x_1,\ldots, x_T$ such that $\ve{x_t}_2\le R_x$ for all $1\le t\le T$.
Then 
there is $k=O\pa{
\ln T \ln \pf{\tau R_\Te R_\Psi R_1 R_xT}{\ep}
}$, $P=O\pf{\tau R_\Te^2 R_\Psi R_1R_xT^{\fc 32}}{\ep}$, 
$M,N\in \R^{W\times k\times n\times m}$, $\be\in \R^\tau$ and $P_0,\ldots, P_{\tau-1}\in \R^{m\times n}$ satisfying
\begin{align}\label{eq:arma-wf-cons1}
\ve{\be}_1 &\le R_1
\\
\label{eq:arma-wf-cons2}
\ve{P_j}_F &\le R_1 R_\Te^2 R_\Psi\\
\label{eq:arma-wf-cons3}
\sqrt{\ve{M}_{2,1}^2 + \ve{N}_{2,1}^2}&=O(R_\Te^2R_\Psi R_1\tau k^{\rc 2})
\end{align}
such that 
the pseudo-LDS $(M,N,\be, P)$ approximates $y_t$ within $\ep$ for $1\le t\le T$:
\begin{align}
\ve{y(M,N,\be,P_{0:\tau-1}; x_{1:t}, y_{t-1:t-\tau})-y_t}&\le \ep.
\end{align}
\end{theorem}

\begin{proof} %[Proof of Theorem~\ref{thm:arma-wf}]
Our plan is as follows. First we show that choosing $\be_j$ and $P_j$ as in Theorem~\ref{thm:arma}, 
\begin{align}
\de_t &= y_t + \sum_{j=1}^{\tau} \be_j  y_{t-j} - \sum_{j=0}^{\tau-1} P_j x_{t-j} 
\end{align}
 can be approximated by 
\begin{align}
\de_t^{(1)} :&= \sum_{\ell=1}^d 
 \sum_{h=1}^k
M_\ell'(h,:,:) \si_h^{\rc 4}(\phi_h^{(\om_\ell)}*x)_{t-\tau}
\end{align}
for $M_\om'\in \R^{k\times n\times m}$; this is obtained by a projection to the wavefilters. Next we approximate $\de_{t}^{(1)}$ by discretizing the phase,
\begin{align}
\de_t^{(2)} :&= \sum_{\ell=1}^{d}
 \sum_{h=1}^k
M_\ell'(h,:,:) \si_h^{\rc 4}(\phi_h^{(e^{2\pi i p_\ell/W})}*x)_{t-\tau}
\end{align}
for some integers $p_\ell\in [0,W-1]$. 
Finally, we show taking the real part gives something in the form
\begin{align}
\sum_{\ell=1}^{d}
\sum_{h=1}^k %\sum_{h=1}^k
%\sum_{u=1}^{t-\tau}
 (M_\ell(h,:,:)\si_h^{\rc 4} (\phi_h^{(\cos, \fc{2\pi p_\ell}{W})}*
x)_{t-\tau}
+(N_\ell(h,:,:)\si_h^{\rc 4} (\phi_h^{(\sin, \fc{2\pi p_\ell}{W})}*
x)_{t-\tau})
\label{eq:real-part}
\end{align}
This matches the form of a pseudo-LDS given in~\eqref{eq:pseudo-LDS} after collecting terms, 
\begin{align}
M(p,:,:,:) &= \sum_{\ell:p_\ell=p} M_\ell&
N(p,:,:,:) &= \sum_{\ell:p_\ell=p} N_\ell.
\end{align}

Now we carry out the plan. Again we have~\eqref{eq:char-poly-y} except that this time the second term is not 0. Let $A=\Psi\La \Psi^{-1}$, $v_\ell$ be the columns of $\Psi$ and $w_{\ell}^*$ be the rows of $\Psi^{-1}$.
By assumption on $p$, $\sumz j{\tau} \be_j \om_\ell^{\tau-j}=0$ for each $\ell$, so the second term of~\eqref{eq:char-poly-y} equals
\begin{align}
\de_t &= C\sum_{i=\tau}^{t-1} \pa{\sum_{\ell=1}^d \sumz j{\tau} \be_j  \al_\ell^{i-\tau}\al_\ell^{\tau-j} v_\ell w_\ell^*}Bx_{t-i} \\
&=
C\sum_{i=\tau}^{t-1} \pa{\sum_{\ell=1}^d \al_\ell^{i-\tau}\sumz j{\tau} \be_j  (\al_\ell^{\tau-j} - \om_\ell^{\tau-j}) v_\ell w_\ell^*}Bx_{t-i}\\
&=\sum_{\ell=1}^dCv_\ell w_\ell^*B \sum_{i=\tau}^{t-1} r_\ell^{i-\tau}\om_\ell^{i-\tau} \sumz j{\tau}  \be_j \om_\ell^{\tau-j} (r_\ell^{\tau-j}-1) x_{t-i}\\
&= \sum_{\ell=1}^dCv_\ell w_\ell^*B \sumz j{\tau}  \sum_{i=\tau}^{t-1} \be_j  \pa{r_\ell^{\tau-j}-1} r_\ell^{i-\tau}
\om_\ell^{i-j}x_{t-i}\\
&=\sum_{\ell=1}^dCv_\ell w_\ell^*B  \sumz j{\tau}  \be_j  \om^{\tau-j} \fc{r^{\tau-j}-1}{1-r} \pa{\mu(r_\ell)^{(\om_\ell)}*x}_{t-\tau}.
\end{align}

Thus this equals
\begin{align}
%&=\sum_{\ell=1}^dCv_\ell\ot w_\ell^*B \sumz j{\tau}  \be_j  \pa{\fc{1-r^{\tau-j}}{1-r}\mu_1(r_\ell)*x^{(\om_\ell)}}_{t-\tau}\\
\de_t &=
\ub{\sum_{\ell=1}^dCv_\ell w_\ell^*B \sumz j{\tau}  \be_j  \om_\ell^{\tau-j} \fc{r^{\tau-j}-1}{1-r} \pa{\wt \mu(r_\ell)^{(\om_\ell)}*x}_{t-\tau}}{\de_t^{(1)}}
\label{eq:arma-wave-term1}
\\
&\quad +
{\sum_{\ell=1}^dCv_\ell w_\ell^*B \sumz j{\tau}    \be_j  \om_\ell^{\tau-j} \fc{r^{\tau-j}-1}{1-r} \pa{(\mu(r_\ell)-\wt \mu(r_\ell))^{(\om_\ell)}*x}_{t-\tau}}%{\ep_t^{(1)}}\label{eq:arma-wave-error}
\end{align}
where $\wt \mu_k(r)$ is the projection of $\mu_k(r)$ to the subspace spanned by the rows of $\phi_1,\ldots, \phi_k$. %, i.e., 
Let $\Phi$ be the matrix with rows $\phi_j$, let $D=\diag((\si_h^{\rc 4})_{1\le h\le k})$ and let $m_\ell\in \R^k$ be such that $\wt \mu(r_\ell) = \Phi Dm_\ell = \sum_{h=1}^k m_{\ell,h}\si_h^{\rc 4}\phi_h$, i.e., $m_\ell = \si_h^{-\rc 4} \an{\phi_h,\mu(r_\ell)}$. The purpose of $D$ is to scale down $\phi_h*x$ so that gradients will be small in the optimization analysis.

We show that the term~\eqref{eq:arma-wave-term1} equals $\de_t^{(1)}$ with some choice of $M_\ell'$. Indeed,
\begin{align}
\eqref{eq:arma-wave-term1}
&= \sumo{\ell}d Cv_\ell w_\ell^* B \sumz j{\tau-1} \be_j  \om_\ell^{\tau-j} \fc{r^{\tau-j}-1}{1-r} \sum_{h=1}^k m_{\ell,h}\si_h^{\rc 4}(\phi_{h}^{(\om_\ell)}* x)_{t-\tau}\\
&=\sumo{\ell}{d}
 \sum_{h=1}^k
 M_\ell'(h,:,:) \si_h^{\rc 4}(\phi_h^{(\om_\ell)} * x)_{t-\tau}
=\de_t^{(1)}\\
\text{where }
M_\ell' (h,r,i):&=(L_\ell)_{ri} m_{\ell,h}\\
L_\ell&= Cv_\ell w_\ell^*B \sumz j{\tau-1} \be_j  \om_\ell^{\tau-j} \pf{r^{\tau-j}-1}{1-r} 
\end{align}
We calculate the error. Here $\ve{x}_\iy$ denotes $\max_i \ve{x_i}_2$.
\begin{align}
\sumo{\ell}d \ve{L_\ell}_F &=\sumo{\ell}d
\ve{Cv_\ell w_\ell^* B\sumz j{\tau-1} \be_j \om_\ell^{\tau-j}
\fc{r^{\tau-j}-1}{1-r}
}_F\\
&\le \sumo{\ell}d \ve{C}_F\ve{B}_F \ve{v_\ell}_2\ve{w_\ell}_2\ve{\be}_1\tau\\
&\le \ve{C}_F\ve{B}_F \sqrt{\sumo{\ell}d\ve{v_\ell}_2^2\sumo{\ell}d\ve{w_\ell}_2^2}\ve{\be}_1\tau
\label{eq:RPsi}\\
&\le R_\Te^2R_\Psi R_1\tau
\label{eq:L-bound}
\\
\ve{\de_t-\de_t^{(1)}}_2 & \le
\sumo{\ell}d \ve{L_\ell}_2
\max_\ell ((\mu(r_\ell) - \wt \mu(r_\ell))^{(\om_\ell)} * x)_{t-\tau}\\
&\le R_\Te^2R_\Psi R_1\tau
\max_\ell \ve{\mu(r_\ell)-\wt \mu(r_\ell)}_1\ve{x}_{\iy}\\ %can actually improve to \sqrt{\tau}
&\le R_\Te^2R_\Psi R_1\tau \sqrt{T}\max_\ell\ve{\mu(r_\ell)-\wt \mu(r_\ell)}_2R_x\\
&\le R_\Te^2R_\Psi R_1\tau \sqrt T c_1^{-k/\ln T}(\ln T)^{\rc 4}.
\end{align}
for some constant $c_1<1$, where the bound on $\ve{\mu(r_\ell)-\wt \mu(r_\ell)}$ comes from Lemma C.2 in \cite{HKZ17}. Choose $k=C\ln T \ln \pf{\tau R_\Te R_\Psi R_1R_x T}{\ep}$ for large enough $C$ makes this $\le \eph$. 

Now we analyze the effect of discretization. Write $\om_\ell = e^{\te_\ell i}$ where $\te_\ell \in [0,2\pi)$, and let $p_\ell$ be such that 
$\ab{\te_\ell - \fc{2\pi p_\ell}{W}}\le \fc{\pi}{W}$ where the angles are compared modulo $2\pi$. 
Let $\om_\ell'=e^{\fc{2\pi p_\ell i}{W}}$. 
We approximate $\de_t^{(1)}$ with 
\begin{align}
\de_t^{(2)} &= \sumo{\ell}d M_\ell'(h,:,:)\si_h^{\rc 4} (\phi_h^{(\om_\ell')} * x)_{t-\tau}
\end{align}
Note that $|\om_\ell^j - \om_\ell'^j|\le j\ab{\te_\ell - \fc{2\pi p_\ell}W}\le \fc{\pi j}{W}$.
We calculate the error. First note
\begin{align}
\sum_{\ell=1}^d \sumo hk \ve{M_\ell'(h,:,:)}_F\si_h^{\rc 4}
&\le \sumo{\ell}d \ve{L_\ell}_2 \max_{\ell,h} |\an{\phi_h,\mu(r_\ell)}| \\
&\le R_\Te^2R_\Psi R_1\tau
\end{align}
by \eqref{eq:L-bound} and $\ve{\mu(r_\ell)}_2\le 1$. Then
\begin{align}
\ve{\de_t^{(1)} - \de_t^{(2)}}_2 
&\le \sumo hk \ve{\sumo{\ell}d M_\ell'(h,i,:) ((\phi_h^{(\om_\ell)} - \phi_h^{(\om_\ell')})*x)_{t-\tau}(i)}_2\\
&\le \pa{\sum_{\ell=1}^d \sumo hk \ve{M_\ell'(h,:,:)}_F\si_h^{\rc 4}}
\max_{h,\ell} \ve{\phi_h^{(\om_\ell)}-\phi_h^{(\om_\ell')}}_1\ve{x}_\iy\\
&\le R_\Te^2R_\Psi R_1\tau \fc{\pi T}{W}\max_h \ve{\phi_h}_1 R_x\\
&\le R_\Te^2R_\Psi R_1\tau \fc{\pi T}{W} \sqrt{T} R_x\\
&= \fc{R_\Te^2R_\Psi R_1\tau\pi T^{\fc 32} R_x}{W}\le\eph
\end{align}
for $W=\fc{2\pi \tau R_\Te^2 R_\Psi R_1R_x T^{\fc 32}}{\ep}$. This means $\ve{\de_t-\de_t^{(2)}}\le \ep$.

Finally, we take the real part of $ \de_t^{(2)}$ (which doesn't increase the error, because $\de_t$ is real)
to get 
\begin{align}
%\sumo uT \phi_j(u) x_{t-u}(i) M_\ell(r,i,j) \om_\ell^u
& \quad 
\sumo{\ell}d
\sum_{i=1}^n \sum_{h=1}^k
\sumo uT \si_h^{\rc 4}\phi_h(u) x_{t-\tau-u}(i) \Re [M_{\ell}'(h,:,i){\om_\ell'}^u]\\
&= \sumo{\ell}d 
\sum_{i=1}^n \sum_{h=1}^k
\sumo uT \si_h^{\rc 4}\phi_h(u) x_{t-\tau - u}(i)
\pa{\Re(M_{\ell}'(h,:,i)) \cos\pf{2\pi p_\ell u}{W} - 
\Im(M_{\ell}'(h,:,i)) \sin\pf{2\pi p_\ell u}{W}
}
%&=\sumo{\ell}d\sumo uT \phi_j(u) x_{t-u}(i)
%|M_{\om_\ell'}(r,i,j)| (\cos(\te_{\ell,r,i,j}) \cos(u\te_\ell) - \sin(\te_{\ell,r,i,j}) \sin(u\te_\ell))\\
\end{align}
which is in the form~\eqref{eq:real-part} with $M_\ell = \Re(M_\ell')$ and $N_\ell =-\Im(M_\ell')$.

The bound on $\ve{M}_{2,1}$ and $\ve{N}_{2,1}$ is
\begin{align}
\sqrt{\ve{M}_{2,1}^2 + \ve{N}_{2,1}^2}& \le \sumo{\ell}d \sqrt{\ve{M_\ell}_F^2 + \ve{N_\ell}_F^2}= \sumo{\ell}d \ve{M_\ell'}_F \\
&=\sumo{\ell}d \sqrt{\sumo hk \ve{M_\ell'(h,:,:)}_F^2}\\
&=\sumo{\ell}d \ve{L_\ell}_F \ve{m_\ell}_2\\
&= O(R_\Te^2R_\Psi^2R_1\tau k^{\rc 2})
\end{align}
because \begin{align}
\ve{m_\ell}_2 =\pa{\sumo hk \si_h^{-\rc 2} \an{\phi_h, \mu(r_\ell)}^2}^{\rc 2}\le\pa{\sumo hk O(1)}^{\rc 2}=O(k^{\rc 2})
\end{align}
by Lemma E.4 in~\cite{HKZ17}.
\end{proof}

\section{Proof of the regret bound}
\label{section:appendix-regret-proof}

In this section, we prove the following theorem:
\begin{theorem}\label{thm:regret}
Let $y^*_1, \ldots, y^*_T$ denote the predictions made by the fixed pseudo-LDS which has the smallest total squared-norm error in hindsight. Then, there is a choice of parameters for which the decision set $\mathcal{K}$ contains all LDSs which obey the assumptions from Section 2.1,
and Algorithm~\ref{alg:asym-lds} makes predictions $\hat y_t$ such that
\[ \sum_{t=1}^T \norm{\hat y_t - y_t}^2 - \sum_{t=1}^T \norm{\hat y_t^* - y_t}^2 \leq
\tilde O\pa{R_1^3 R_x^2 R_\Theta^4 R_\Psi^2 R_y^2 d^{5/2} n \log^7 T \sqrt{T}}, \]
where the $\tilde O(\cdot)$ only suppresses factors polylogarithmic in $n,m,d,R_\Theta,R_x,R_y$.
\end{theorem}

\subsection{Online learning with composite strongly convex regularizers}
\label{subsection:multi-strongly-convex}

Algorithm~\ref{alg:asym-lds} runs the follow-the-regularized-leader (FTRL) algorithm with the regularization
\[ R(M,N,\beta,P) := \norm{M}^2_{2,q} + \norm{N}^2_{2,q} + \norm{\beta}_{q'}^2 + \sum_{j=1}^\tau \norm{P_j}_F^2, \]
where $q = \frac{\ln(W)}{\ln(W)-1}$ and $q' = \frac{\ln(\tau)}{\ln(\tau)-1}$.
To achieve the desired regret bound, we need to show that this regularizer is strongly convex with respect to the composite norm considered in the algorithm. We will work with the following definition of strong convexity with respect to a norm:
\begin{definition}[Strong convexity w.r.t. a norm]
A differentiable convex function $f : \mathcal{K} \rightarrow \reals$ is $\alpha$-strongly convex with respect to the norm $\norm{\cdot}$ if, for all $x, x+h \in \mathcal{K}$, it holds that
\[ f(x+h) \geq f(x) + \langle \nabla f(x),h\rangle + \frac{\alpha}{2} \norm{h}^2. \]
\end{definition}
We first verify the following claim:
\begin{lemma}
\label{lem:multi-strongly-convex-norm}
Suppose the convex functions $R_1, \ldots, R_n$, defined on domains $\mathcal{X}_1, \ldots, \mathcal{X}_n$, are $\alpha$-strongly convex with respect to the norms $\norm{\cdot}_1, \ldots, \norm{\cdot}_n$, respectively. Then, the function $R(x_1, \ldots, x_n) = \sum_{i=1}^n R_i(x_i)$, defined on the Cartesian product of domains, is $(\alpha/n)$-strongly convex w.r.t. the norm $\norm{(x_1, \ldots, x_n)} = \sum_{i=1}^n \norm{x_i}_i$.
\end{lemma}
\begin{proof}
Summing the definitions of strong convexity, we get
\begin{align*}
R(x_1 + h_1, \ldots, x_n + h_n) &\geq \sum_{i=1}^n \langle{ \nabla R(x_i), h_i \rangle} + \fc{\al}{2}\sum_{i=1}^n \norm{h_i}_i^2 \\
&= \langle \nabla R(x_1, \ldots, x_n), \mathrm{vec}(h_{1:n}) \rangle
+ \frac{\alpha}{2} \sum_{i=1}^n \norm{h_i}_i^2 \\
&\geq \langle \nabla R(x_1, \ldots, x_n), \mathrm{vec}(h_{1:n}) \rangle+\frac{\alpha}{2n} \pa{ \sum_{i=1}^n \norm{h_i}_i }^2,
\end{align*}
where the last inequality uses the AM-QM inequality.
\end{proof}

Indeed, each term in $R(\hat\Theta)$ is strongly convex in the respective term in the composite norm $\norm{\hat\Theta}$. For $\norm{M}_{2,q}^2$ (and, identically, $\norm{N}_{2,q}^2$), we use Corollary 13 from \cite{kakade2012regularization}:
\begin{lemma}
The function $M \mapsto \norm{M}^2_{2,q}$ is $\frac{1}{3 \ln(W)}$-strongly convex w.r.t. the norm $\norm{\cdot}_{2,1}$.
\end{lemma}
This is a more general case of the fact that $\beta \mapsto \norm{\beta}_{q'}^2$ is $\frac{1}{3\ln(\tau)}$-strongly convex w.r.t. the norm $\norm{\beta}_1$.
Finally, $\sum_{j=1}^\tau \norm{P_j}_F^2$ is the squared Euclidean norm of
$\mathrm{vec}(P_{1:\tau})$, which is clearly 1-strongly convex w.r.t. the same Euclidean norm. Thus, applying Lemma~\ref{lem:multi-strongly-convex-norm}, we have:
\begin{corollary}
$R(\hat \Theta)$ is $\alpha$-strongly convex in $\norm{\hat\Theta}$, where
$\alpha = \frac{1}{12 \ln(\max(\tau,P))}.$
\end{corollary}

Finally, we note an elementary upper bound for the dual of the norm $\norm{\hat \Theta}$ in terms of the duals of the summands, which follows from the definition of dual norm:
\begin{lemma}
For any norm $\norm{\cdot}$, let $\norm{\cdot}^*$ denote its dual norm $\norm{v}^* = \sup_{\norm{w} \leq 1} \langle v,w \rangle$. Then, if $\norm{(x_1, \ldots, x_n)} = \sum_{i=1}^n \norm{x_i}_i$, we have
\[ \norm{(x_1, \ldots, x_n)}^* \leq \sum_{i=1}^n \norm{x_i}_i^*. \]
\end{lemma}
\begin{corollary}
In particular, for the norm $\norm{\hat \Theta}$ we have defined on pseudo-LDSs, we have:
\[ \norm{ (M,N,\beta,P) }^* \leq \norm{M}_{2,\infty} + \norm{N}_{2,\infty} + \norm{\beta}_\infty + \sqrt{ \sum_{j=1}^\tau \norm{P_j}_F }. \]
\end{corollary}

\subsection{Regret of the FTRL algorithm}

We state the standard regret bound of FTRL with a regularizer $R$ which is strongly convex with respect to an arbitrary norm $\norm{\cdot}$. For a reference and proof, see Theorem 2.15 in \cite{shalev2012online}.
\begin{lemma}
In the standard online convex optimization setting with decision set $\mathcal{K}$, with convex loss functions $f_1, \ldots, f_T$, let $x_{1:t}$ denote the decisions made by the FTRL algorithm, which plays an arbitrary $x_1$, then $x_{t+1} := \argmin_x \sum_{u=1}^t f_t(x) + \frac{R(x)}{\eta}$. Then, if $R(x)$ is $\alpha$-strongly convex w.r.t. the norm $\norm{\cdot}$, we have the regret bound
\[ \mathsf{Regret} := \sum_{t=1}^T f_t(x_t) - \min_x \sum_{t=1}^T f_t(x) \leq \frac{2 R_\mathrm{max}}{\eta} - \frac{\eta T}{\alpha} G_\mathrm{max}^2, \]
where $R_\mathrm{max} = \sup_{x \in \mathcal{K}} R(x)$
and $G_\mathrm{max} = \sup_{x \in \mathcal{K}} \norm{\nabla f(x)}^*$.
\end{lemma}
To optimize the bound, choose $\eta = \fc{\sqrt{2 R_\mathrm{max} T / \alpha}}{G_\mathrm{max}} $, for a regret bound of
\[ \mathsf{Regret} \leq O\pa{ G_\mathrm{max} \sqrt{R_\mathrm{max} T / \alpha} }. \]

With the facts established in Section~\ref{subsection:multi-strongly-convex}, this gives us the following regret bound, which gives an additive guarantee versus the best pseudo-LDS in hindsight:
\begin{corollary}
\label{cor:alg-rftl-regret-bound}
For the sequence of squared-loss functions on the predictions $f_1, \ldots, f_T : \mathcal{K} \rightarrow \reals$, Algorithm 1
produces a sequence of pseudo-LDSs $\hat \Theta_1, \ldots, \hat \Theta_T$ such that
\[ \sum_{t=1}^T f_t( \hat\Theta_t ) - \min_{\hat\Theta \in \mathcal{K}} \sum_{t=1}^T f_t( \hat\Theta ) \leq O \left( G R_{\hat\Theta} \sqrt{T \log \max(P,\tau)} \right), \]
where $G$ is an upper bound on the quantity
\[\norm{\nabla_M f_t(\hat\Theta)}_{2,\infty} + \norm{\nabla_N f_t(\hat\Theta)}_{2,\infty} + \norm{\nabla_\beta f_t(\hat\Theta)}_\infty
+ \sum_{j=1}^\tau \norm{\nabla_P f_t(\hat\Theta)}_F.\]
(Here, $\nabla_M$ denotes the 4-tensor of partial derivatives with respect to the entries of $M$, and so on.)
\end{corollary}
It suffices to establish an upper bound $R_{\hat\Theta}$ on the norm of a pseudo-LDS required to approximate a true LDS, as well as the gradient of the loss function in each of these dual norms. We can obtain the appropriate diameter constraint from Theorem~\ref{thm:arma-wf}:
\[R_{\hat\Theta} = R_1 + R_1 R_\Theta^2 \sqrt{\tau} + 2 R_\Theta^2 R_\Psi^2 R_1 \tau \sqrt{k} = \Theta\pa{R_\Theta^2 R_\Psi R_1 \tau \sqrt{k}}.\]
We bound $G$ in the following section.

\subsection{Bounding the gradient}
In this section, we compute each gradient, and bound its appropriate norm.

\begin{lemma}
\label{lem:grad-bound}
Let $G$ be the bound in the statement in \ref{cor:alg-rftl-regret-bound}. Then,
\[ G \leq O\left( R_1^2 R_x^2 R_\Theta^2 R_\Psi R_y^2 \tau^{3/2} nk^{3/2} \log^2 T \right). \]
\end{lemma}

\begin{proof}
First, we use the result from Lemma~E.5 from \cite{HKZ17}, which states that the $\ell_1$ norm of $\phi_h$ is bounded by $O(\log T / \sigma_h^{1/4})$, to bound the size of the convolutions taken by the algorithm:
\begin{lemma}
\label{lem:filter-l1}
For any $h \in [1,k]$ and $p \in [0, W-1]$, we have
\[ \norm{ \sum_{u=1}^T \phi_h(u) \sigma_h^{1/4} \cos(2\pi u p / W) x_{t-u} }_2 \leq O\pa{ R_x \sqrt{n} \log T }. \]
The same holds upon replacing $\cos(\cdot)$ with $\sin(\cdot)$.
\end{lemma}
\begin{proof}
We have that for each coordinate $i \in [n]$,
\begin{align}
| \sum_{u=1}^T \phi_h(u) \sigma_h^{1/4} \cos(2\pi u p / W) x_{t-u}(i) | \\
\leq \sum_{u=1}^T \phi_h(u) \sigma_h^{1/4} R_x
\leq \norm{\phi_h}_1 \sigma_h^{1/4} R_x \\
\leq O\pa{ R_x \log T }.
\end{align}
\end{proof}

It will be useful to record an upper bound for the norm of the prediction residual $y(\hat\Theta) -  y_t$, which appears in the gradient of the least-squares loss.
By assumption, we have $\norm{y_t}_2 \leq R_y$.
By the constraint on $\norm{\Theta}$ from the algorithm, and noting that $y(\hat \Theta)$ a sum of matrix products of $M(p,h,:,:)$ and convolutions of the form of the LHS in Lemma~\ref{lem:filter-l1}, we can obtain a bound on $y(\hat\Theta)$ as well:
\begin{align}
\norm{ y(\hat\Theta) - y_t }_2 &\leq O\pa{ R_y \norm{ y(\hat\Theta) }_2 } \\
&\leq O\pa{R_y R_{\hat\Theta} \cdot (\sqrt{nk} \log T R_x + R_x + R_y) } \\
&\leq O \pa{ R_1^2 R_x R_\Theta^2 R_\Psi R_y^2 \tau \sqrt{n}k \log T }.
\end{align}

Call this upper bound $U$ for short.
First, we compute the gradients with respect to the 4-tensors $M$ and $N$. Fixing one phase $p$ and filter index $k$, we have:
\begin{align}
\nabla_M f_t(\hat \Theta) (p,h,:,:) = 2 \pa{ y(\hat\Theta) - y_t }\pa{ \sum_{u=1}^T \phi_h(u) \cos(2\pi u p / W) x_{t-u} }^\top,
\end{align}
so that
\begin{align}
\norm{ \nabla_M f_t(\hat \Theta) (p,h,:,:) }_F^2 &\leq 4 \norm{y(\hat\Theta) - y_t}^2 \norm{ \sum_{u=1}^T \phi_h(u) \cos(2\pi u p / W) x_{t-u} }^2 \\
&\leq U \cdot O\pa{ R_x \sqrt{n} \log T }
\end{align}
Thus, we have
\begin{align}
\norm{ \nabla_M f_t(\hat \Theta) }_{2,\infty} &= \max_{p} \sqrt{ \sum_{h=1}^k \norm{ \nabla_M f_t(\hat \Theta) (p,h,:,:) }_F^2 } \\
&\leq U \cdot O\pa{ R_x \sqrt{nk} \log T }
\end{align}
The same bound holds for $\norm{ \nabla_N f_t(\hat \Theta) }_{2,\infty}$.

For the $\beta$ part of the gradient, we have
\begin{align}
\nabla_\beta f_t(\hat \Theta) (j) = 2 \pa{ y(\hat\Theta) - y_t }^\top y_{t-j},
\end{align}
so that we have an entrywise bound of
\begin{align}
\norm{ \nabla_\beta f_t(\hat \Theta) }_\infty \leq O\pa{ R_y \cdot U }.
\end{align}

Finally, for the $P_j$ part, we have
\begin{align}
\nabla_{P_j} f_t(\hat \Theta) = 2 \pa{ y(\hat\Theta) - y_t } x_{t-j}^\top,
\end{align}
so that
\begin{align}
\sqrt{ \sum_{j=1}^\tau \norm{ \nabla_{P_j} f_t(\hat \Theta) }_F^2 } \leq O \pa{ \sqrt{\tau} R_x \cdot U }.
\end{align}

The claimed bound on $G$ follows by adding these bounds.
\end{proof}

The final regret bound follows by combining Lemma~\ref{lem:grad-bound} and Corollary~\ref{cor:alg-rftl-regret-bound}, with the choices $k = \Theta\pa{ \log T \log \pf{\tau R_\Te R_\Psi R_1R_x T}{\ep} }$, $\tau = \Theta(d), W = \Theta\pa{ \tau R_\Te^2 R_\Psi R_1R_x T^3 }$.

\section{Proof of main theorem: Competitive ratio bounds}
\label{app:main}
\subsection{Perturbation analysis}
To prove the main theorem, we first need to analyze the approximation when there is noise. Compared to the noiseless case, as analyzed in Theorem~\ref{thm:arma-wf-precise}, we incur an additional term equal to the size of the perturbation times a {\it competitive ratio} depending on the dynamical system.

\begin{lemma}\label{lem:perturb}
Consider an LDS $\Te=(A,B,C,h_0=0)$ that satisfies the conditions of Theorem~\ref{thm:arma-wf}. 
%and let $(M,N, \be,P)$ be chosen as in the theorem.
Consider the LDS under adversarial noise~\eqref{eq:lds-noise1}--\eqref{eq:lds-noise2}. Let $y_{t}(x_{1:T}, \eta_{1:T},\xi_{1:T})$ be the output at time $t$ given inputs $x_{1:T}$ and noise $\eta_{1:T},\xi_{1:T}$.  Let $\hat y_t(x_{1:T},\eta_{1:T},\xi_{1:T})$ be the prediction made by the pseudo-LDS at time $t$. 
%Given inputs $x_1,\ldots, x_T$, let the outputs be $y_1^*,\ldots, y_T^*$. 
Suppose that $(M,N,-\be,P)$ is a pseudo-LDS that predicts well when there is no noise: %if $\hat y_t$ is the prediction at time $t$,
\begin{align}
\ve{\hat y_{1:T}(x_{1:T},0,0) - y_{1:T}(x_{1:T},0,0)}_2:&=\pa{\sumo tT \ve{\hat y_t(x_{1:T},0,0)  - y_t^*(x_{1:T},0,0) }_2^2}^{\rc 2}\le \ep.
\end{align}
%Now consider the same LDS but under adversarial noise~\eqref{eq:lds-noise1}--\eqref{eq:lds-noise2} satisfying $\sumo tT \ve{\eta_t}^2 + \ve{\xi_t}^2\le L$. Given inputs $x_1,\ldots, x_T$, let $y_1,\ldots, y_T$ be the outputs of this system. 
Then for bounded adversarial noise $\sumo tT \ve{\eta_t}^2 + \ve{\xi_t}^2\le L$,
\begin{align}
\ve{\hat y_{1:T}(x_{1:T},\eta_{1:T},\xi_{1:T}) - y_{1:T}(x_{1:T},\eta_{1:T},\xi_{1:T})}_2 &\le \ep + O(\ve{\be}_\iy \tau^{\fc 32} R_\Te R_\Psi \sqrt L).
%\sumo tT \ve{y(M,N,\be,P_{0:\tau-1}, x_{1:t},y_{t-1:t-\tau}) - y_t}^2\le (\ep+\ve{\be}_\iy \tau^{\fc 32} R_\Te L)^2.
\end{align}
\end{lemma}
Note that the initial hidden state can be dealt with by considering it as noise in the first step $\eta_1$.
\begin{proof}
Note that $\hat y_t$ is a linear function of $x_{1:T},\eta_{1:T}, \xi_{1:T}$ so 
\begin{align}
\hat y_t(x_{1:T},\eta_{1:T},\xi_{1:T})-y_t &= [\hat y_t(0,0,\xi_{1:T}) - y_t(0,0,\xi_{1:T})]
 + [\hat y_t(0,\eta_{1:T},0) - y_t(0,\eta_{1:T},0)] \\
 &\quad + [\hat y_t(x_{1:T}, 0,0) - y_t(x_{1:T}, 0,0) ].\label{eq:perturb0}
\end{align}
%Consider the error $\sqrt{\sumo tT\ve{\hat y_t - y_t^*}^2}$. Because the predictions $\hat y_t$ respond linearly to $\eta_t, \xi_t$, this error is at most the sum of the errors incurred by each $\eta_t$ and $\xi_t$ individually.
This says that the residual is the sum of the residuals incurred by each $\xi_t$ and $\eta_t$ individually, plus the residual for the non-noisy LDS, $\hat y_t(x_{1:T}, 0,0) - y_t(x_{1:T}, 0,0)$.

We first analyze the effect of a single perturbation to the observation $\xi_t$. Suppose $\eta_t=0$ for all $t$ and $\xi_t=0$ except for $t=u$ (so that $y_t=0$ for all $t$ except $t=u$, where $y_u=\xi_u$). The predictions $\hat y_t$ are zero when $t\nin [u, u+\tau]$, because then the prediction does not depend on $y_u$. For $u\le t\le u+\tau$,
%\hl{be consistent with sign of $\be$}
\begin{align}
\ve{\hat y_t(0,0,\xi_{1:T}) - y_t(0,0,\xi_{1:T})}_2&=
\ve{y_t + \sumo j{\tau}\be_j y_{t-j}}_2
%&= \max_{0\le j\le \tau}\ve{\be_j}_2\\
\le |\be_{t-u}| \ve{\xi_{u}}_2\\
\ve{\hat y_{1:T}(0,0,\xi_{1:T})-y_{1:T}(0,0,\xi_{1:T})}_2
&\le\pa{\sum_{t=u}^{u+\tau} |\be_{t-u}|^2}^{\rc 2} \ve{\xi_u}_2
%\sqrt{\sumo tT\ve{y_t - \hat y_t}_2^2}&
=\ve{\be}_2\ve{\xi_u}_2.\label{eq:perturb1}
\end{align}

Now we analyze the effect of a single perturbation to the hidden state $\eta_t$. Suppose $\xi_t=0$ for all $t$ and $\eta_t=0$ except for $t=u$ (so that $h_t=0$ for all $t<u$, $h_u=\eta_u$, and the system thereafter evolves according to the LDS). 
For simplicity, we may as well consider the case where $u=1$, i.e., the perturbation is to the initial hidden state.
When $t\le \tau$, the error is bounded by 
%When $u\le t< u+\tau$, the error is bounded by
\begin{align}
\ve{\hat y_t(0,\eta_{1:T},0)- y_t(0,\eta_{1:T},0)}_2&=
\ve{y_t + \sumo j{t-1}\be_j y_{t-j}}_2
= \ve{C \sumz j{t-1}\be_j A^{t-j-1}\xi_1}\\
&\le \ve{C}_2 \ve{\be}_1 \ve{\Psi}_F \ve{\La}_2 \ve{\Psi^{-1}}_F\ve{\eta_1}_2
\le R_\Te R_\Psi \ve{\be}_1 \ve{\eta_1}_2\\
\ve{\hat y_{1:\tau}(0,\eta_{1:T},0)-\hat y_{1:\tau}(0,\eta_{1:T},0)}_2
&\le \tau^{\rc 2} R_\Te R_\Psi \ve{\be}_1 \ve{\eta_1}_2.\label{eq:perturb2}
\end{align}
When $t>\tau$, %u+\tau
the error is (let $\be_0=1$)
\begin{align}
%\ve{y_t - \hat y_t}_2
y_t(0,\eta_{1:T},0)-
\hat y_t(0,\eta_{1:T},0)
&= {C \sumz j{\tau}\be_j A^{t-j-1}\eta_1}\\
&\le C\sumz{j}{\tau} \sumo{\ell}d \be_j \al_\ell^{t-j-1} v_\ell w_\ell^* \eta_1\\
&= C\sumz{j}{\tau} \sumo{\ell}d \be_j \al_\ell^{t-\tau-1} (\al_\ell^{\tau-j} - \om_\ell^{\tau-j}) v_\ell w_\ell^* \eta_1\\
&= C\sumo{j}{\tau} \sumo{\ell}d \be_j \om_\ell^{t-j}r_\ell^{t-\tau-1} (r_\ell^{\tau-j}-1) v_\ell w_\ell^* \eta_1\\
\ve{\hat y_t(0,\eta_t,0)-y_t(0,\eta_{1:T},0)}_2
&\le \ve{\be}_\iy \sumo j{\tau}\sumo{\ell}d 
\fc{1-r_\ell^{\tau-j}}{1-r_\ell}\mu(r_\ell)_{t-\tau} \ve{C v_\ell w_\ell^* \eta_1}_2\\
&\le  \ve{\be}_\iy  R_\Te \ve{\eta_1}_2\sumo j{\tau}\sumo{\ell}d 
\fc{1-r^{\tau-j}}{1-r}\mu(r_\ell)_{t-\tau}\ve{v_\ell w_\ell^* }_2\\
\ve{\hat y_{\tau+1:T}(0,\eta_{1:T},0)-y_{\tau+1:T}(0,\eta_{1:T},0)}_2
&\le \ve{\be}_\iy R_\Te \ve{\eta_1}_2 \sumo j{\tau} \max_{r,\ell} \ve{\fc{1-r_\ell^{\tau-j}}{1-r_\ell}\mu(r_\ell)}_2\sumo{\ell}d\ve{v_\ell w_\ell^*}_2\\
&\le \ve{\be}_\iy R_\Te \ve{\eta_1}_2 \pa{\sumo j{\tau} \sqrt{\tau-j}}R_\Psi\le \ve{\be}_\iy R_\Te R_\Psi \tau^{\fc 32}\ve{\eta_1}_2\label{eq:perturb3}
\end{align}
by the calculation in~\eqref{eq:RPsi} and~\eqref{eq:L-bound} because 
\begin{align}
\ve{\fc{1-r^{k}}{1-r}\mu(r)}_2
&\le (1-r^k)\sfc{1}{1-r^2}\le 
\min\{1,k(1-r)\} \sfc{1}{1-r}\\
&=\min\bc{\sfc{1}{1-r}, k\sqrt{1-r}}\le \sqrt k.
\end{align}
%The total error $\hat y_t-y_t$ 
%is the sum of the errors accrued by the $\eta_t$ and $\xi_t$, plus the error in approximating the noiseless system, $\hat y_t-y_t^*$.
Combining~\eqref{eq:perturb1},~\eqref{eq:perturb2}, and~\eqref{eq:perturb3} using~\eqref{eq:perturb0} and noting $\ve{\be}_1\le(\tau+1)\ve{\be}_\iy$ gives
\begin{align}
\ve{\hat y_{1:T}(x_{1:T}, \eta_{1:T},\xi_{1:T})-y_t}_2 &\le 
\ve{\be}_2 \ve{\eta}_2
+
(\tau^{\rc2} R_\Te R_\Psi \ve{\be}_1 + \ve{\be}_\iy R_\Te R_\Psi \tau^{\fc 32}) \ve{\xi}_2 + \ep\\
&\le O(\ve{\be}_\iy\tau^{\fc 32} R_\Te R_\Psi \sqrt{L})+\ep
%\ve{\be}_\iy R_\Te\tau^{\fc 32}\ve{\eta}_2
%+\ep \le \ve{\be}_\iy R_\Te \tau^{\fc 32} L+\ep.
\end{align}
\end{proof}
\subsection{Proof of Main Theorem}
\label{app:main-pf}
We prove Theorem~\ref{thm:main} with the following more precise bounds.

\begin{theorem}\label{thm:main-formal}
Consider a LDS with noise satisfying the assumptions in Section~\ref{sec:setting} (given by~\eqref{eq:lds-noise1} and~\eqref{eq:lds-noise2}), where total noise is bounded by $L$. 
Then there is a choice of parameters such that
%with parameters $k=O\pa{\poly\log\pa{T, R_\Te, R_\Psi, R_1, R_x, \rc{\ep}}}$, $P=O\pa{\poly(\tau, R_\Te, R_\Psi, R_1, R_x, T)}$, 
%\hl{what is the regularization weight?} 
Algorithm~\ref{alg:asym-lds} learns a pseudo-LDS $\hat \Te$ whose predictions $\hat y_t$ satisfy
\begin{align}\label{eq:main-formal}
\sumo{t}T \ve{\hat y_t - y_t}^2 &\le 
\tilde O\pa{R_1^3 R_x^2 R_\Theta^4 R_\Psi^2 R_y^2 d^{5/2} n \log^7 T \sqrt{T}} + 
O(R_\iy^2 \tau^{3} R_\Te^2 R_\Psi^2 L)
\end{align}
where the $\tilde O(\cdot)$ only suppresses factors polylogarithmic in $n,m,d,R_\Theta,R_x,R_y$.
\end{theorem}
\begin{proof}
Consider the fixed pseudo-LDS which has smallest total squared-norm error in hindsight for the noiseless LDS. Let $y_t^*(x_{1:T},\eta_{1:T},\xi_{1:T})$ denote its predictions under inputs $x_{1:T}$ and noise $\eta_{1:T},\xi_{1:T}$ and $y_t(x_{1:T},\eta_{1:T},\xi_{1:T})$ denote the true outputs of the system.
Given $\ep>0$, choosing $k,P$ as in  Theorem~\ref{thm:arma-wf-precise} (Approximation) gives
\begin{align}
\forall 1<t<T,\quad \ve{y_t^*(x_{1:T},0,0)-
y_t(x_{1:T},0,0)}_2
&\le \ep\\
\implies  \ve{y_{1:T}^*(x_{1:T},0,0)-
y_{1:T}(x_{1:T},0,0)}_2
&\le \ep\sqrt T.
\end{align}
By Lemma~\ref{lem:perturb} (Perturbation),
\begin{align}
\ve{y_{1:T}^*(x_{1:T},\eta_{1:T},\xi_{1:T})-
y_t(x_{1:T},\eta_{1:T},\xi_{1:T})}_2
&\le \ep \sqrt T + O(R_\iy \tau^{\fc 32} R_\Te R_\Psi \sqrt L)\label{eq:main-perturb}
\end{align}

%Let $y_1^*,\ldots, y_T^*$ denote the predictions made by the fixed pseudo-LDS which has the smallest total squared-norm error in hindsight. We use the notation $\ve{a_{1:T}}^2:= \sumo tT a_t^2$ as in Lemma~\ref{lem:perturb}. 
In the below we consider the noisy LDS (under inputs $x_{1:T}$ and noise $\eta_{1:T},\xi_{1:T}$). 
By Theorem~\ref{thm:regret} (Regret) and~\eqref{eq:main-perturb},
\begin{align}
\ve{\hat y_{1:T} - y_{1:T}}_2^2
&= (
\ve{\hat y_{1:T} - y_{1:T}}_2^2 - 
\ve{\hat y_{1:T}^* - \hat y_{1:T}}_2^2 
)
+\ve{\hat y_{1:T}^* - y_{1:T}}_2^2\\
&\le 
\tilde
O\pa{R_1^3 R_x^2 R_\Theta^4 R_\Psi^2 R_y^2 d^{5/2} n \log^7 T \sqrt{T}} + 
 \ep^2 T + O(R_\iy^2 \tau^{3} R_\Te^2 R_\Psi^2 L)
\end{align}
Choosing $\ep={T^{-\rc 4}}$ (and $k,P$ based on $\ep$) means $\ep^2T$ is absorbed into the first term, and we obtain the bound in the theorem.
\end{proof}

\end{document}